\documentclass[runningheads]{llncs}
\usepackage{graphicx}

\PassOptionsToPackage{hyphens}{url}\usepackage[breaklinks=true,bookmarks=false]{hyperref}
\usepackage{tikz}
\usepackage{nicefrac, xfrac}
\usepackage{mathtools}
\usepackage{booktabs}
\usepackage{multirow}
\usepackage{enumitem}

\usepackage{fancyhdr}

\newcommand\graytext[1]{\textcolor{gray}{{#1}}}

\begin{document}
\definecolor{solarized}{RGB}{253,246,227}
\title{Fusing Hand and Body Skeletons for \\ Human Action Recognition in Assembly}
\titlerunning{Fusing Hand and Body Skeletons for Human Action Recognition}
\author{Dustin Aganian \and
        Mona Köhler \and
        Benedict Stephan \and
        Markus Eisenbach \and
        Horst-Michael Gross
        \thanks{\scriptsize{This work has received funding from the Carl-Zeiss-Stiftung as part of the project engineering for smart manufacturing (E4SM).}}}
\authorrunning{D. Aganian et al.}
\institute{Ilmenau University of Technology, Neuroinformatics and Cognitive Robotics Lab 98693 Ilmenau, Germany\\
\email{dustin.aganian@tu-ilmenau.de, ORCID: 0009-0006-3925-6718}}
\maketitle              %

\newboolean{isarxiv}
\setboolean{isarxiv}{true}
\ifthenelse{\boolean{isarxiv}}{%
    \renewcommand{\headrulewidth}{0pt}
    \addtolength{\headwidth}{2cm}
    \fancypagestyle{fancyfirstpage}{%
        \fancyhf{}%
        \fancyhead[C]{%
            \scriptsize%
             \hspace{-2cm}
               \textcolor{lightgray}{%
                  \small%
                   This work has been submitted to ICANN 2023 for possible publication.\\
                   \hspace{-2.2cm}
                   Copyright may be transferred without notice, after which this version may no longer be accessible.%
               }%
        }
        \fancyfoot[C]{%
            \footnotesize%
            \textcolor{gray}{\thepage}%
        }
    }
    \fancypagestyle{fancypage}{%
        \fancyhf{}%
        \fancyfoot[C]{%
            \footnotesize%
            \textcolor{gray}{\thepage}%
        }
    }
    \thispagestyle{fancyfirstpage}
    \pagestyle{fancypage}
}{%
    \thispagestyle{empty}%
    \pagestyle{empty}%
}%

\begin{abstract}
As collaborative robots (cobots) continue to gain popularity in industrial manufacturing, effective human-robot collaboration becomes crucial.
Cobots should be able to recognize human actions to assist with assembly tasks and act autonomously.
To achieve this, skeleton-based approaches are often used due to their ability to generalize across various people and environments. 
Although body skeleton approaches are widely used for action recognition, they may not be accurate enough for assembly actions where the worker's fingers and hands play a significant role.
To address this limitation, we propose a method in which less detailed body skeletons are combined with highly detailed hand skeletons.
We investigate CNNs and transformers, the latter of which are particularly adept at extracting and combining important information from both skeleton types using attention.
This paper demonstrates the effectiveness of our proposed approach in enhancing action recognition in assembly scenarios.
\keywords{Action Recognition \and Skeleton-based \and Fusion \and Body Skeletons \and Hand Skeletons \and 3D/2D Skeletons \and Assembly \and Deep Learning.}
\end{abstract}

\section{Introduction}
Collaborative robots are playing an increasingly important role in the course of Industry 4.0~\cite{inkulu2021challenges}.
In order for the robot to collaborate with a human worker and assist in assembly processes, it first needs to visually perceive its environment, the current assembly state, and human actions~\cite{eisenbach2021,terreran2023skeleton,wang2019symbiotic}.
For human action recognition, often RGB-based approaches are utilized in the state of the art, as they achieve the best results. %
However, RGB-based approaches face major difficulties when the target scenario deviates from the training scenario.
They tend to overfit to the environment and the persons seen, especially when the training dataset lacks diversity~\cite{terreran2023skeleton}.
This limitation frequently applies to assembly datasets~\cite{attach,IKEA-wacv2021}, which are often small and recorded at only a few locations.
In contrast, skeleton-based approaches do not face these limitations, as they only process skeletons and, thus, can generalize much better to different environments.

\begin{figure}[!t]
    \centering
    \includegraphics[width=0.99\hsize]{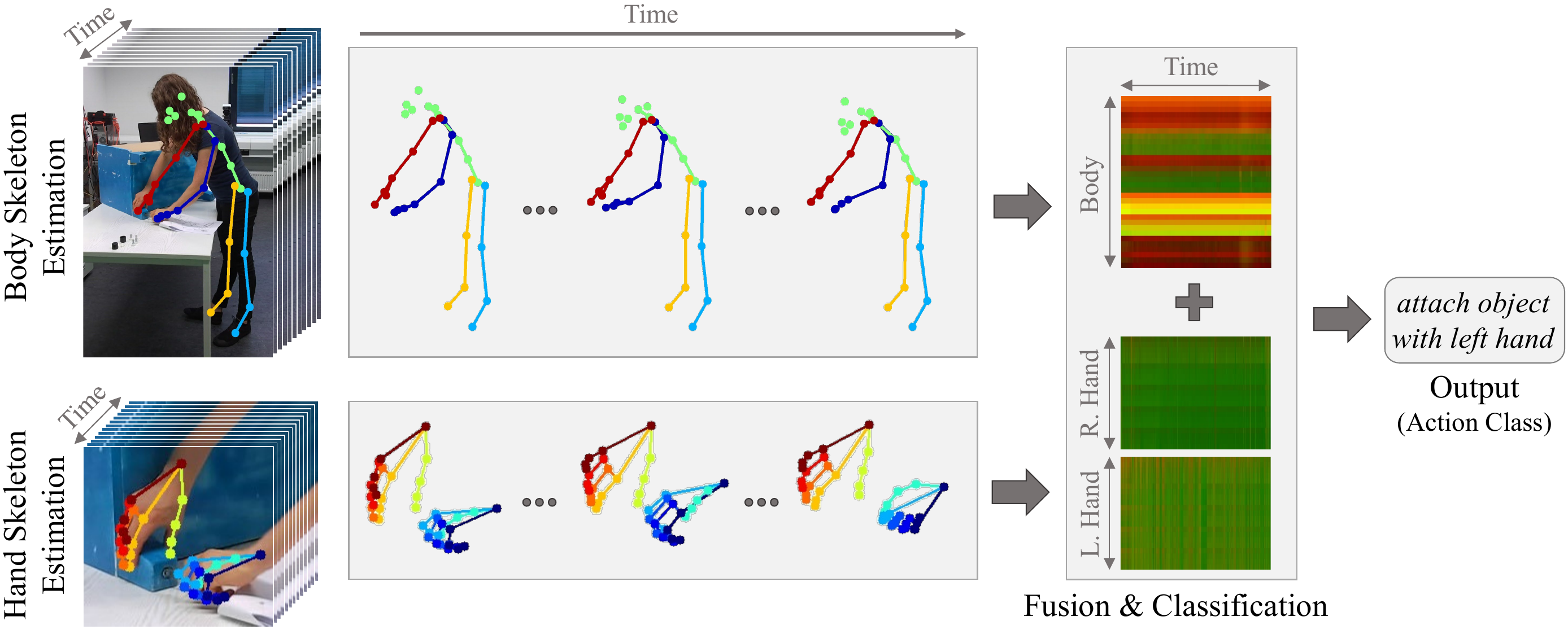}
    \vspace*{-4mm}
    \caption{We combine body skeletons with hand skeletons for human action recognition. 
    Some actions can be recognized primarily by the movement of the hands.
    The encoding of the skeleton sequences to images is explained in \autoref{sec:approach_baseline}.
    Example frames from~\cite{attach}.
    }
    \vspace*{-6mm}
    \label{fig:overview}
\end{figure}

However, as shown in \autoref{fig:overview}, some actions are difficult to recognize by the body skeleton alone.
For example, the action of attaching a small object is mainly characterized by the object movement, as utilized in \cite{aganian2023IJCNN}, and  how the worker's hands interact with it. 
For this assembly step, it is therefore also useful to utilize finer hand skeletons.
This is already done for other assembly datasets~\cite{meccano,assembly101}, which are recorded in first-person view.
However, using hand skeletons alone might not be sufficient for actions such as turning, rotating or pushing of workpieces.
During these actions, the fingers are mostly rigid, and most of the movement takes place in the upper body.

Therefore, in this paper, we want to investigate how highly detailed hand skeletons can be combined with less detailed body skeletons to enhance the recognition of assembly actions on the ATTACH~\cite{attach} and IKEA ASM~\cite{IKEA-wacv2021} datasets.
By doing so, we aim to recognize both types of actions.

Our study examines both 2D and 3D body skeletons.
While 3D skeletons offer a more comprehensive representation of the person's actions, 2D skeletons are more widely available in practical applications.
In this paper, we demonstrate how 2D and 3D hand skeletons can be integrated with various body skeletons.
One of the key challenges is that the hands are often occluded, either partially or entirely, which can complicate the estimation of hand positions and the fusion with body skeletons.
We also explore the challenges associated with differently detailed skeletons. %
Specifically, a body skeleton typically has 18-32 joints, while two hand skeletons have most often 42 joints.
Although, there are typically more hand joints than body joints, the latter contains significantly more crucial information for many assembly actions.
Therefore, in this paper, we describe how to address this dimension imbalance.
Our contributions are as follows:
\begin{enumerate}[nosep]
    \item We investigate the use of hands in conjunction with body skeletons in both 2D and 3D to improve action recognition for assembly tasks.
    \item We predict hand skeletons on the ATTACH and the IKEA ASM datasets and employ a selection process to identify the appropriate hands.
    \item To the best of our knowledge, we are the first to employ the SwinV2 transformer~\cite{swinv2-cvpr2022} for skeleton-based action recognition.
\end{enumerate}

\section{Related Work}
\label{sec:sota}
In the following, we first present the state of the art of action recognition with skeleton sequences, before going into more detail about differences between hand and body skeleton action recognition and possibilities of fusing skeletons.

\subsection{Methods for skeleton-based Action Recognition}
Human action recognition encompasses various subfields, but in this paper, we focus on the action classification task of pre-trimmed video clips of human skeleton sequences, as this task serves as a foundation for other related problems, such as action segmentation or action detection.
For skeleton-based action recognition, recently, 2D convolutional neural networks (CNNs) such as VA-CNN~\cite{VACNN-TPAMI2019}, 3D CNNs like PoseConv3D~\cite{PoseConv3D-cvpr2022}, graph convolution networks (GCNs) such as 2S-AGCN~\cite{Shi2019AGCN}, and transformers like AcT~\cite{mazzia2022action} have been used.

In our paper, we adopt the skeleton encoding of~\cite{skeleton-encoding-acpr2015} and use it like VA-CNN, which employs a ResNet50~\cite{ResNet-cvpr2016} backbone, as it has demonstrated superior or comparable results on the ATTACH Dataset~\cite{attach} compared to GCN methods.
In this approach, the skeleton sequence is encoded as an image so that typical image based classifiers can be used.
The image encoding also provides the ability to weight the different skeletons based on their occupied image space which will be explained in \autoref{sec:approach}.

Moreover, we are able to replace the CNN backbone with the SwinV2-T transformer~\cite{swinv2-cvpr2022}, which has demonstrated excellent results in image-based pattern recognition.

\subsection{Hand and Body Skeleton-based Action Recognition}
The idea of fusing less detailed body skeletons with highly detailed hand skeletons for action recognition has only been briefly addressed in the literature, and is still a new area of research.
For instance, in NTU-X~\cite{trivedi2021ntu} body skeletons from NTU-RGBD 60/120 were extended to include highly detailed hand skeletons and facial features.
In~\cite{trivedi2023psumnet} a model was trained for every skeleton type to build an ensemble for classifying actions.
It was demonstrated in~\cite{trivedi2021ntu,trivedi2023psumnet} that additional hand skeletons from the NTU dataset for everyday and domestic actions (such as making gestures, eating or blowing one's nose) are helpful to the classification task.

In contrast, during assembly, the hand is often occluded by the object being worked on, and the quality of the estimated hand skeletons varies significantly.
Typically, the state of the art for action recognition with hands focuses on gesture recognition, where the hands are usually unoccluded.
For action recognition during assembly, hand skeletons have only been used in fine motor assembly (e.g., Meccano~\cite{meccano}, Assembly101~\cite{assembly101}), where cameras are mounted either on the worker's head or above the table and focus on the worker's arms and hands.
For instance, in the application scenario of fine-motor toy assembly, which is similar to ours,~\cite{assembly101} demonstrated that estimated hand skeletons can be utilized for action recognition.
This indicates that our approach of fusing body skeletons with hand skeletons shows promise for action recognition in general assembly tasks.
Such tasks involve a combination of coarse actions, where the movement of the body is relevant, and fine motor actions (as in \autoref{fig:overview}), where hand skeletons are primarily important.
Therefore, in this paper, our goal is to explore how these differently detailed body and hand skeletons can be combined optimally. %

\section{Hand and Body Skeleton Dataset Preparation}
\vspace*{-1.5mm}

Below, we first present the datasets we used.
Afterwards, we explain how we estimated the hand skeletons and what to consider when processing them.

\subsection{Datasets}
\label{sec:datasets}
\vspace*{-1.5mm}

To show our approach, we utilize two datasets that contain both small-grained assembly actions that can be mainly recognized by the hands movement as well as coarse assembly actions that involve the whole body, namely the ATTACH~\cite{attach} and the IKEA ASM~\cite{IKEA-wacv2021} datasets.
Both datasets are captured from multiple views as shown in \autoref{fig:datasets} and consist of assembly actions, where IKEA furniture are assembled.
The action names for the action recognition task are composed of verb-object pairs.
Below, we discuss each dataset and its characteristics in detail.

\vspace*{-1mm}
\subsubsection{ATTACH}
Overall, the ATTACH dataset~\cite{attach} comprises 51.6 hours of recordings from 42 persons with about 95K action instances for action recognition distributed over 51 classes.
While there are different training splits available, we specifically focus on the person split in this paper as it is the most commonly used split for action recognition. 

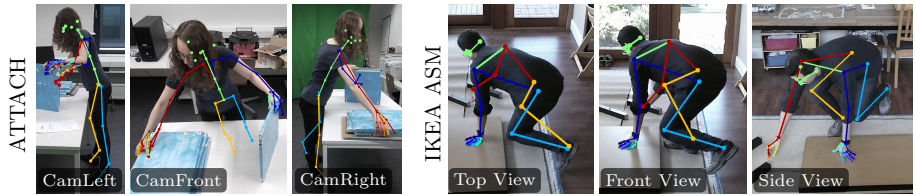
\begin{figure}[!t]
    \centering
    \resizebox{\linewidth}{!}{%
    \begin{tikzpicture}%
        \tikzstyle{camera_label}=[anchor=north east, text=white, fill=black, fill opacity=0.6, text opacity=1, inner sep=2pt, rounded corners=2pt, minimum height=3.5mm, font=\scriptsize]
        \node[anchor=north west, rotate=90] at (-0.375, -2.35){\footnotesize{ATTACH}};%
        \node[anchor=north west] at (0, 0){\includegraphics[height=2.7cm, trim=460 150 60 100, clip]{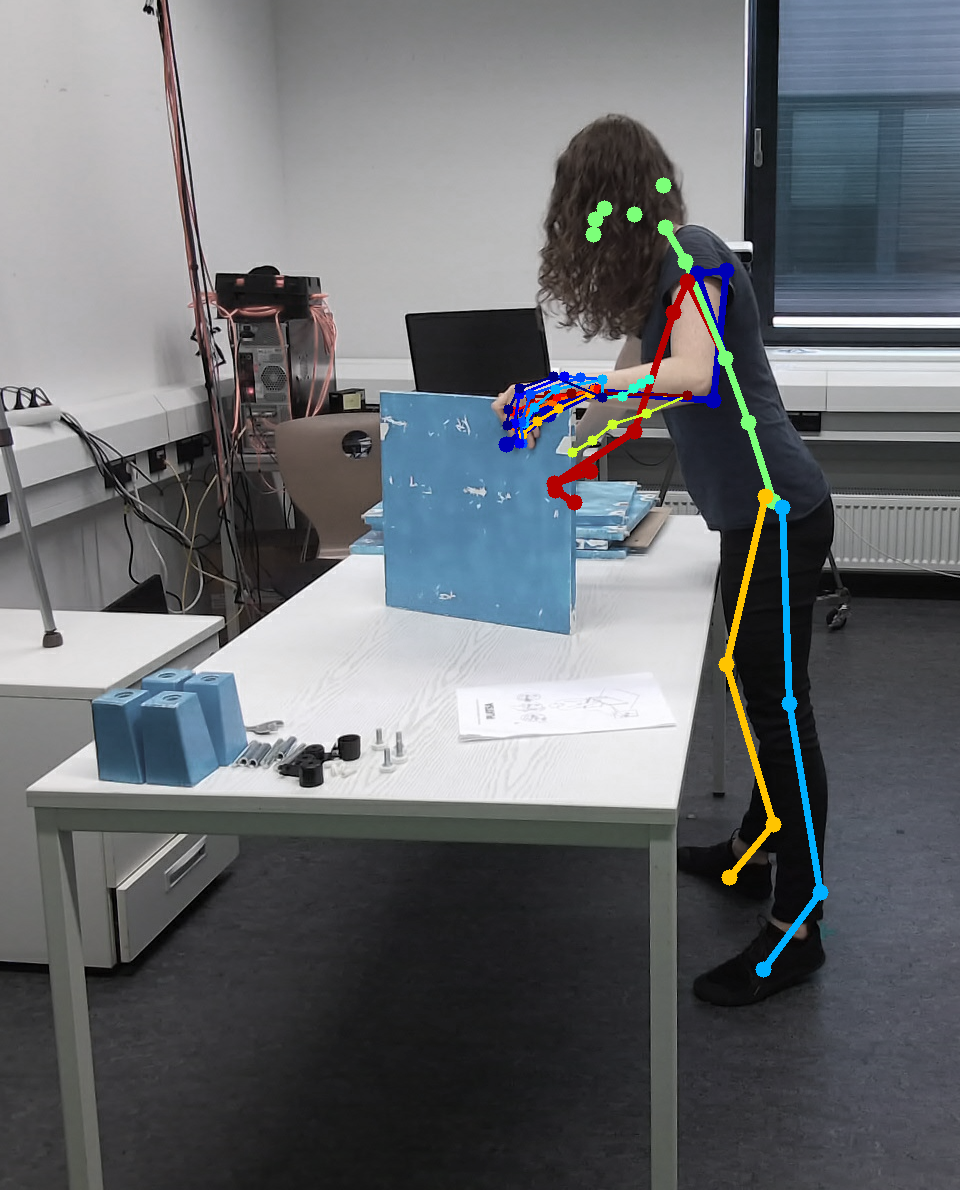}};%
        \node[camera_label] at (1.35, -2.45){CamLeft};%
        \node[anchor=north west] at (1.35, 0){\includegraphics[height=2.7cm, trim=200 150 700 150, clip]{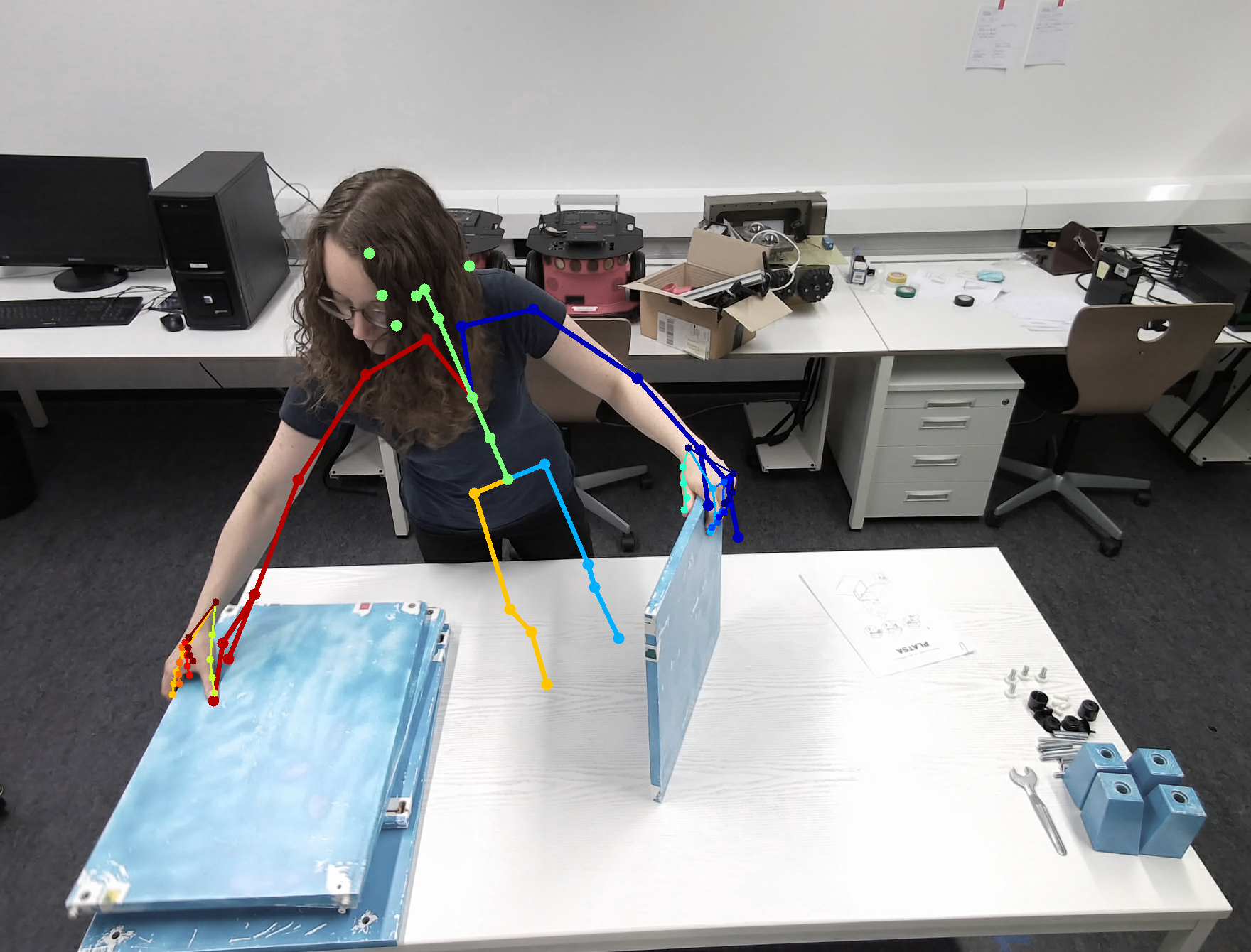}};%
        \node[camera_label] at (2.85, -2.45){CamFront};%
        \node[anchor=north west] at (3.67, 0){\includegraphics[height=2.7cm, trim=190 30 380 30, clip]{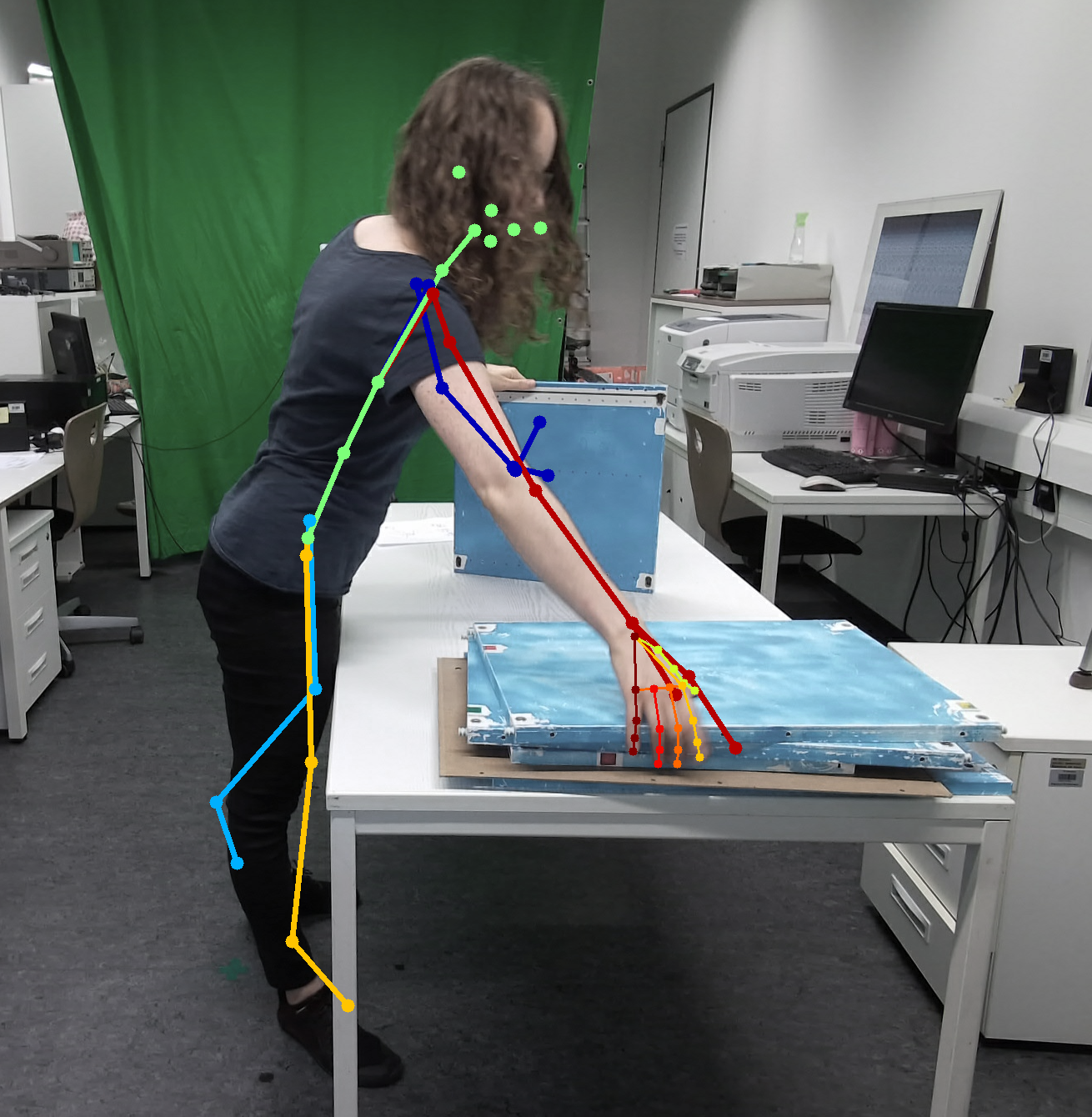}};%
        \node[camera_label] at (5.22, -2.45){CamRight};%
        \node[anchor=north west, rotate=90] at (5.575, -2.48){\footnotesize{IKEA ASM}};%
        \node[anchor=north west] at (5.9, 0){\includegraphics[height=2.7cm, trim=310 250 40 50, clip]{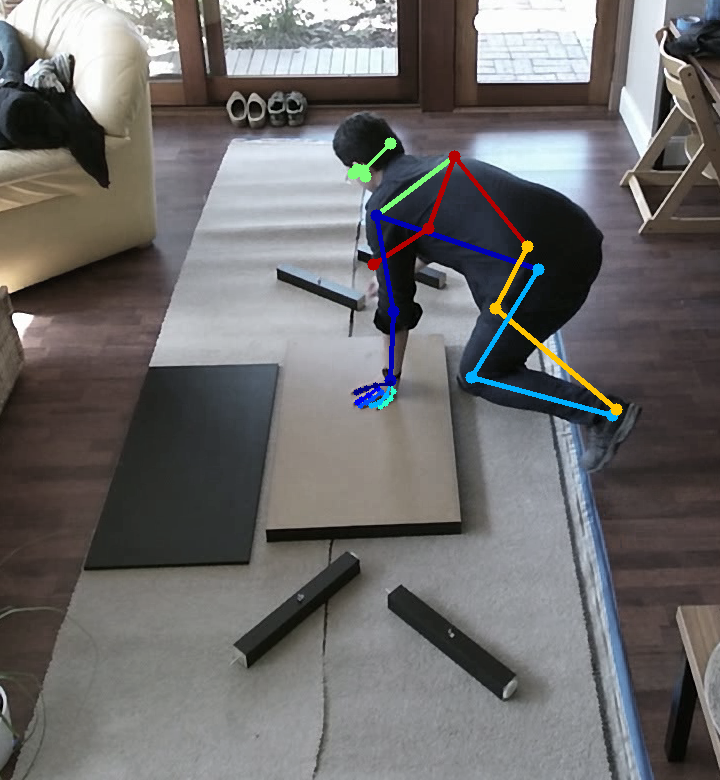}};%
        \node[camera_label] at (7.35, -2.45){Top View};%
        \node[anchor=north west] at (8.07, 0){\includegraphics[height=2.7cm, trim=360 200 20 80, clip]{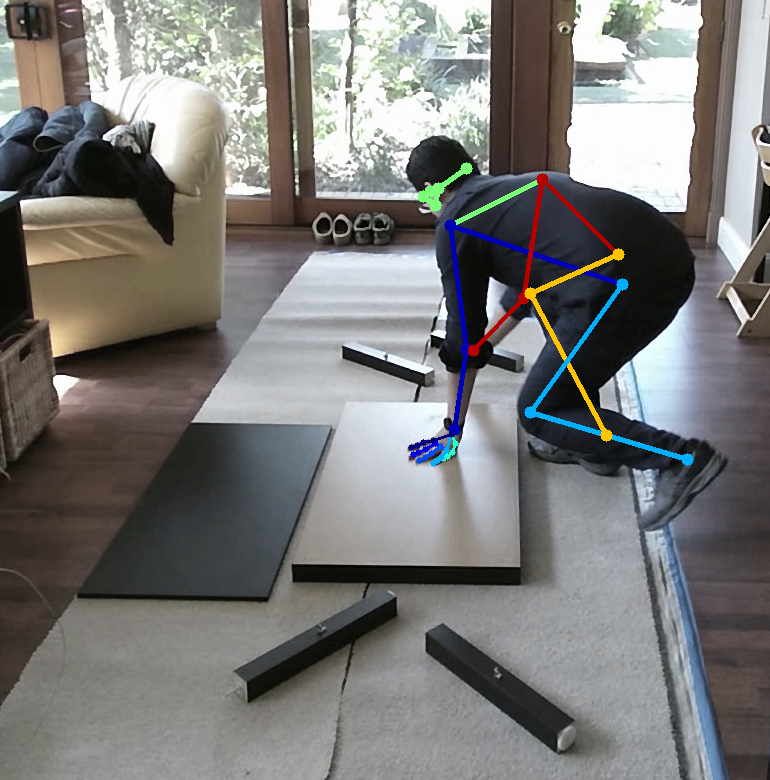}};%
        \node[camera_label] at (9.72, -2.45){Front View};%
        \node[anchor=north west] at (10.26, 0){\includegraphics[height=2.7cm, trim=160 140 350 140, clip]{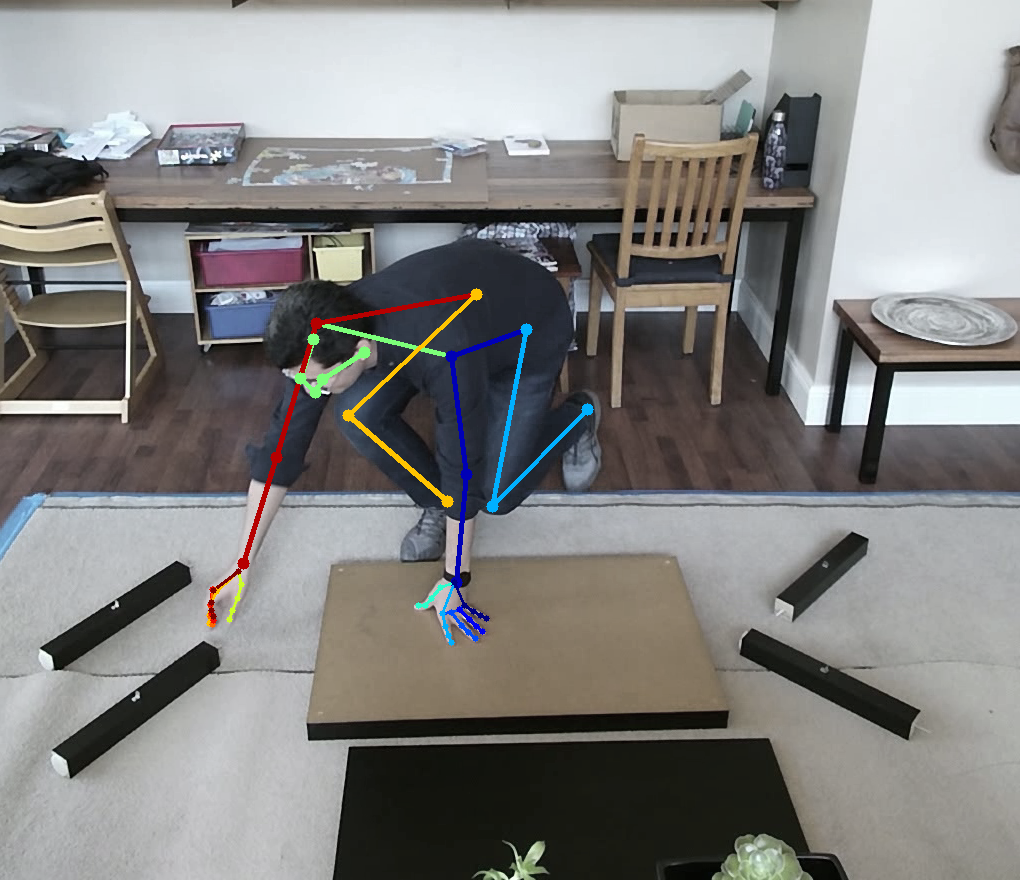}};%
        \node[camera_label] at (11.76, -2.45){Side View};%

    \end{tikzpicture}%
    }
    \vspace*{-3mm}
    \caption{%
    Cropped example images of the ATTACH~\cite{attach} and IKEA ASM~\cite{IKEA-wacv2021} datasets, overlayed with body and hand skeletons.
    The body skeletons originate from the datasets; the hand skeletons are estimated with MediaPipe~\cite{mediapipe}.
    }
    \label{fig:datasets}
\end{figure}

Skeleton data are available in 3D from the Azure Kinect framework.
The skeletons are composed of 32 joints and are located in a metric space with the origin in the camera.
Since the state of the art typically deals with 2D skeletons, we have also transformed the 3D skeletons into the 2D frame of the RGB camera. %
In our experiments, we consider both 3D and 2D body skeletons for combination with hand skeletons.
As can be seen in \autoref{fig:datasets} (left), the Kinect Azure body skeleton already contains four rudimentary joints for each hand (wrist, palm and tips of index finger and thumb).

It is worth noting that actions are labeled for each hand independently.
Moreover, some actions involve the use of tools such as wrenches, hammers or screwdrivers, where most of the movement occurs in the hand and fingers.
Intuitively, this suggests that incorporating additional hand skeletons could potentially enhance the performance of skeleton-based action recognition methods.

\vspace*{-1mm}
\subsubsection{IKEA ASM}
The IKEA ASM dataset~\cite{IKEA-wacv2021} consists of 35.3 hours of recordings from 48 different persons with about 17K action instances distributed over 33 classes.
We use the official splits provided in~\cite{IKEA-wacv2021}. %
The dataset provides 2D skeletons for all views estimated by OpenPose~\cite{openpose-cvpr2017} and Keypoint R-CNN~\cite{maskrcnn-iccv2017}, of which we use the latter as it performed better in preliminary experiments.
Unlike the Kinect skeleton, both OpenPose as well as Keypoint R-CNN only predict one single wrist joint per hand, as shown in \autoref{fig:datasets} (right).
Therefore, incorporating additional hand skeletons might also be useful for action recognition on the IKEA ASM dataset.

However, it should be noted that some actions are difficult to recognize even with hand skeletons.
For example, actions such as \emph{pick up back panel}, \emph{pick up front panel}, and \emph{pick up side panel} can only be distinguished by the object used, which is not present in the skeleton data.

\subsection{Hand Skeleton Estimation}
For estimating hand skeletons, the hands need to be clearly visible in the current frame.
However, due to their small size in the IKEA ASM dataset, we first cropped a $300{\times}300$ patch of the RGB image around the wrist joint of the body skeleton.
For the ATTACH dataset, we can skip this first step.

To detect hands and estimate hand skeletons, we used MediaPipe~\cite{mediapipe}.
However, since the predictions can be rather noisy, we filtered the hands by discarding all hands where the distance between the wrist joints of the predicted hand skeleton and the body skeleton exceeded a certain threshold.
We kept at most two hands per image.
In cases where hand skeletons were missing, we simply took skeletons from past frames to attribute for the missing data.

MediaPipe predicts both 2D and 3D hand skeletons with 21 joints each.
While 2D hand skeletons are represented in the image plane, the 3D hand skeletons are represented in a metric space, where the origin is located on the surface of each hand.
Therefore, when working with 3D data, we transformed the 3D hands into the frame of the 3D body skeletons.
This allows for better capturing the spatial relationships between the movements of the hands and the body.

\section{Approach}
\label{sec:approach}

In the following, we describe our approach to action recognition of pre-trimmed skeleton sequences.
In \autoref{sec:approach_baseline}, we first present our baseline with body skeletons, before discussing different variations for incorporating hand skeletons in \autoref{sec:approach_fusion}.

\begin{figure}[!t]
    \centering
    \includegraphics[width=\hsize]{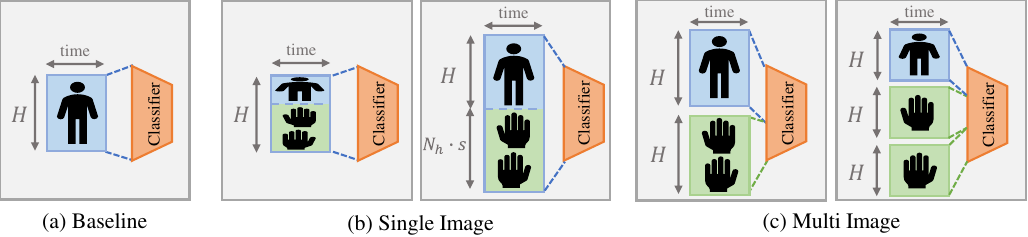}
    \vspace*{-7mm}
    \caption{Overview of our different fusion approaches.
    (a) As a baseline we train models with only the body skeleton.
    $H$ is the height of the input image. %
    (b) As a simple way of fusing both skeleton types we merge them into a single image while investigating different ratios between body and hand skeletons. $N_h$ is the number of hand joints ($42$ in our case) and $s$ is a scaling factor.
    (c) We treat both skeletons types as different modalities and apply them as distinct input images.}
    \label{fig:fusion_approach}
    \vspace*{-4.5mm}
\end{figure}

\subsection{Baseline: Body Skeleton Approach}
\label{sec:approach_baseline}
For our baseline, we only use the body skeleton, without incorporating additional hand skeletons\footnote{\scriptsize{Our preliminary experiments on the ATTACH dataset using hand skeletons solely showed far inferior results compared to body skeletons solely and are thus not investigated further.}}.
For this, we encode the skeletons from a trimmed action sequence into one single RGB image, similar to~\cite{skeleton-encoding-acpr2015,VACNN-TPAMI2019}.
One column of the image represents one frame, where the skeleton joints are stacked in a fixed order.
To transform a joint to RGB, the XYZ coordinates are normalized and scaled.\footnote{\scriptsize{For ResNet, the image is resized to $224{\times}224$ with pixel values ranging from 0-255. For Swin, we use a resolution of $256{\times}256$ with pixel values from 0-1.}}
For 2D skeleton data, we have just two channels.
These images (see \autoref{fig:overview} for a visualization) can then be used as input to typical image-based classification architectures such as ResNet50 (ResNet)~\cite{ResNet-cvpr2016}.

Furthermore, while ResNet is typically used in the state of the art~\cite{attach,VACNN-TPAMI2019}, we additionally use a SwinV2-T transformer (Swin)~\cite{swinv2-cvpr2022} for the first time to classify skeleton sequences.
Moreover, Swin offers another possibility for fusing hand and body skeleton data, which we will describe in the following.

\subsection{Approaches for Fusing Hand and Body Skeletons}
\label{sec:approach_fusion}
For incorporating additional hand skeleton data, we experiment with different methods, as illustrated in \autoref{fig:fusion_approach}.
\autoref{fig:fusion_approach}a serves as a schematic representation of our baseline approach. %
In the following, we describe two approaches for encoding the sequence of body skeletons with additional hand skeletons.
The first approach involves encoding the hand and body skeletons in a single image, while the second approach creates multiple images that are then combined in the network, similar to multimodal networks that integrate color data with depth data~\cite{emsaformer2023ijcnn,esanet2021icra}.

\subsubsection{Single Image Fusion}
\autoref{fig:fusion_approach}b illustrates our single image fusion approach.
Naively, the hand skeleton joints could be appended below the body skeleton joints in the skeleton encoded image.
For example, for a Kinect Azure skeleton and MediaPipe hands, the first 32 rows would contain the body skeleton, followed by the right hand and the left hand, each with 21 rows.
In this way, however, the body skeleton would account for just under 43\% of the input, while the hands would account for the remaining 57\%.
Such a division, in which the number of hand joints of both hands is predominant, is typical for the relevant skeletons used in the state of the art.
This example is shown on the left side of \autoref{fig:fusion_approach}b.

However, for recognizing assembly actions, the body skeleton provides more relevant information than the fine hand skeletons, which should only serve as support.
With such a naive partitioning of the image, the classifier is given a bias by devoting a larger input space to the hand skeletons.

To address this issue, we investigate another option to fuse the skeletons into one image, which is shown on the right side of \autoref{fig:fusion_approach}b.
Here, we keep the original scaling resolution of the body skeleton as in the baseline (see \autoref{fig:fusion_approach}a).
The body skeleton is scaled up to the original input resolution of the classifier, and subsequently, another image with upscaled hand skeletons is stacked below.
We investigate scaling factors $s \in [1,8]$, where we scale the height of the encoded hand skeleton images (i.e., $N_h = 42$ for MediaPipe skeletons), where $s{=}8$ resembles the scaling of the body skeleton image.  %

\vspace*{-3mm}
\subsubsection{Multiple Image Input}
As an alternative to the previous approach, the skeleton data can be split into different images, and the resulting features can be fused in the network.
Recent work on the EMSAFormer~\cite{emsaformer2023ijcnn} has shown that the SwinV2 transformer is particularly suitable for multimodal processing.
In their study, the Swin transformer was extended in such a way that RGB and depth images of a scene are fed into the same Swin network as two different images.

We propose a similar approach for processing the encoded body skeleton images and the encoded hand skeleton images.
\autoref{fig:fusion_approach}c (left) shows how we create two images, one for the body skeleton and one for both hands.
The first image is encoded on the first 64 channels of the feature map in the patch embedding, and the second image is encoded on the last 32 channels.
After the first attention block, the network combines the information and passes it on to the subsequent blocks, whereby the Swin architecture was not changed.

Alternatively, we can split the skeletons into three images, as shown in \autoref{fig:fusion_approach}c (right).
In this case, three images are created, and each is embedded on 32 channels and given to the respective attention head.
With this approach, the network itself can decide how to further use the combined information.

We also attempted to use the ESANet Encoder~\cite{esanet2021icra} approach, in which two parallel executed ResNets exchange feature maps at different stages.
However, as expected, no improvements were found compared to using only the body skeletons.
This is because these approaches look for pixel correspondences in the images, which are not present in the given skeleton encoding.

\section{Experiments}

Below we present the results of our experiments on fusing body skeletons with hand skeletons.
In \autoref{sec:experiments_3d}, we show experiments with 3D body skeletons, before moving on to 2D body skeletons in \autoref{sec:experiments_2d}.
First, we describe our training setup.

\subsection{Setup}
\label{sec:setup}
Our networks were trained for 100 epochs using the Adam optimizer and a one cycle learning rate scheduler with 10\% of epochs as warmup and several maximum learning rates ranging from $5\cdot10^{-3}$ to $5\cdot10^{-5}$. %
We validated after each epoch and chose the best epoch for testing.
The performances of our trained networks are evaluated using mean class accuracy (mAcc) and top-1 accuracy (top1), two widely used metrics in action recognition literature~\cite{attach,meccano,assembly101,trivedi2021ntu}.

Our networks are initialized with ImageNet weights, which improves performance, although the encoded images generated from skeleton data differ a lot from real images.
However, performance is still fluctuating, which is why we trained with at least five well-functioning learning rates and repeated training three times for each setup.
We present our result using box plots, where each box plots summarizes at least 15 trainings.

\subsection{Experiments with 3D Body Skeletons}

In the following, we present results solely on the ATTACH dataset~\cite{attach}.
While the IKEA ASM dataset~\cite{IKEA-wacv2021} also includes 3D skeletons, they are only available for one camera perspective and captured at a very low frame rate, which makes them rather unsuitable for skeleton-based action recognition.

\subsubsection{Baseline -- 3D Body Skeletons}
\label{sec:experiments_3d}

\begin{figure}[!t]
    \centering
    \includegraphics[width=\hsize]{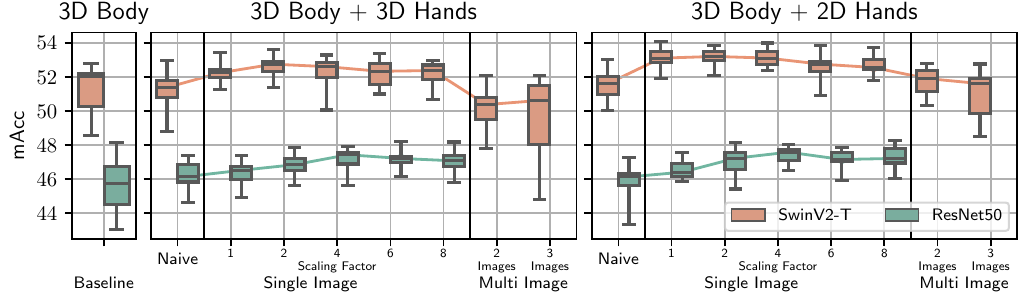}
    \caption{Results using 3D body skeletons on the ATTACH dataset for our baseline models as in \autoref{fig:fusion_approach}a and our different fusion methods: Naive concatenation as in \autoref{fig:fusion_approach}b left, image concatenation with scaling of encoded hand skeleton image as in \autoref{fig:fusion_approach}b right, multi image input as in \autoref{fig:fusion_approach}c. Best results are listed in \autoref{tab:bestResults}.}
    \label{fig:experiments_3d_boxplot}
    \vspace*{-4mm}
\end{figure}

This subsection serves as a benchmark for our experiments with fused inputs, as we optimize hyperparameters to create a strong baseline using body skeletons solely.
On the left side of \autoref{fig:experiments_3d_boxplot}, we present the results of the baseline experiments with 3D body skeletons on the ATTACH dataset.
We compare the performance of two models with similar complexity, namely the SwinV2-T transformer (Swin) and the ResNet50 CNN (ResNet).
Our results demonstrate that Swin outperforms ResNet, with a median improvement of more than six percentage points and a maximum improvement of more than four percentage points.
Even the worst performing Swin model performs better than the best ResNet model, indicating that Swin is a suitable model for processing skeleton sequences encoded as images.

However, we want to emphasize that training with Swin is significantly more challenging than with ResNet, which is usually very robust regarding hyperparameters. %
With Swin, it is crucial to select an appropriate learning rate schedule, as training can fail with even slightly too high learning rates. %
Conversely, slightly too low learning rates do not produce significant improvements over ResNet.
We found that the best results were achieved with learning rates only marginally smaller than the ones that caused training to fail.

\subsubsection{Fusion of Hand Skeletons with 3D Body Skeletons}

\autoref{fig:experiments_3d_boxplot} illustrates the results of our fusion experiments using our transformed 3D hand skeletons from MediaPipe in the middle, and 2D hand skeletons on the right.

\paragraph{3D Hands:}
Overall, an improvement of the median and variance can be observed when using the single image fusion approach with the correct scaling factor.
While no improvement of the maximum for ResNet is observable, for Swin the incorporation of 3D hands increased performance by about one percentage point.
This shows that there is relevant information in the hand skeletons that helps making the training more consistent or even improves the general quality of the models.
Moreover, it shows that Swin is significantly better at combining the relevant information from the estimated hand skeletons with the full body skeletons.
However, since the 3D hand skeletons in MediaPipe are estimated on 2D color images, a poor estimation of the hand joints may have led to only slight improvements.
Therefore, we explore to combine the 3D body skeleton with 2D hand skeletons in the following.

\paragraph{2D Hands:}
The results of fusing 2D hand skeletons with 3D body skeletons are shown in the right half of \autoref{fig:experiments_3d_boxplot}.
First and foremost, this fusion can be challenging due to the different frames of reference.
The 3D skeletons exist in a metric space while the 2D skeletons are given in image coordinates.
This means that the different parts of the input image for the single image fusion approach need to be normalized independently.

For ResNet, using the 2D hands results in similar performance compared to 3D hand skeletons.
On the other hand, Swin demonstrates that this fusion works very well, and in some cases, it performs even better than the fusion with 3D hands.
In fact, the maximum improvement over the baseline is more than one percentage point.
This highlights Swin's ability to handle the challenges of using disparate input spaces.

These results also confirm our assumption that the estimated 3D hand skeletons from MediaPipe are less accurate than the 2D hand skeletons.

\paragraph{Fusion Variations:}
When comparing the different fusion approaches that we examined, both ResNet and Swin yielded similar results.
The naive approach, which involves stacking the hand and body skeleton joints and then scale the encoded skeleton image (\autoref{fig:fusion_approach}b left), produced inferior results compared to stacking the encoded images for hand and body skeleton joints (\autoref{fig:fusion_approach}b right).
This highlights the importance of scaling up the body skeleton image with a higher upscaling factor, similar to the body skeleton baseline (\autoref{fig:fusion_approach}a).

However, we observed different results when comparing how much the hand skeleton joints need to be upscaled.
Swin performed better with a smaller scale factor, while ResNet achieved better results with a larger scale factor.
This could possibly be attributed to the different convolutions in the first layer of the respective networks - ResNet uses a $7{\times}7$ convolution with stride 2, while Swin's patch embedder is a $4{\times}4$ convolution with stride 4.

We also compared single image fusion approaches to multiple image approaches in the Swin transformer.
Unfortunately, the multiple image approaches were inferior to all other approaches.
The median and maximum results were significantly worse, and the variance was much larger.
This suggests that this approach for multimodal input to a Swin network cannot be easily applied. %

The lower performance of the multiple image approaches in Swin could potentially be attributed to the patch embedding process.
This involves splitting the convolutions to different images to obtain the feature maps with the needed channel sizes. 
Furthermore, we experimented with larger patch embeddings as in~\cite{emsaformer2023ijcnn}, where the body skeleton image is processed into 96 channels of the feature map and the hands into 32 or both into 64.
Although this improved the models and made them perform similarly to the single image approaches, it significantly increased the needed computational power and training time.
In~\cite{emsaformer2023ijcnn}, it was shown that appropriate pre-training can be crucial.
However skeleton-based pre-training is not typical in literature and also not the focus of this paper.

\subsection{Experiments with 2D Body Skeletons}
\label{sec:experiments_2d}

Most datasets and state-of-the-art approaches utilize 2D skeletons. %
Therefore, we also experiment with 2D skeletons and show results on the ATTACH~\cite{attach} and the IKEA ASM~\cite{IKEA-wacv2021} dataset.
First, we present results of our body only baseline and afterwards the fusion with hand skeletons.

\subsubsection{Baseline -- 2D Body Skeletons}
\begin{figure}[!t]
    \centering
    \begin{tikzpicture}{
        \node at (0, 2.2) {ATTACH};
        \node at (6, 2.2) {IKEA ASM};
        \node at (0,0){\includegraphics[width=0.49\hsize]{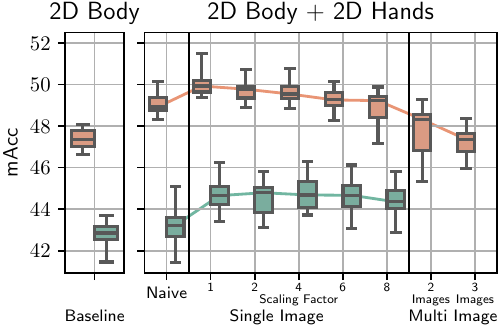}};
        \node at (6, 0){\includegraphics[width=0.49\hsize]{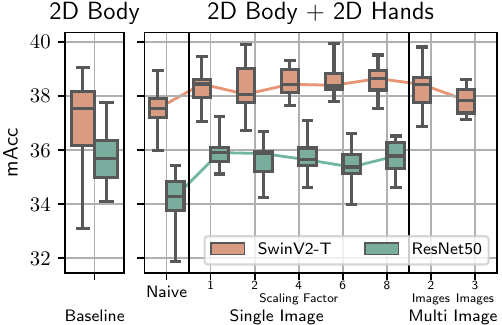}};
    }
    \end{tikzpicture}
    \vspace*{-8mm}
    \caption{Results using 2D body skeletons on the ATTACH and IKEA ASM datasets for our baseline models as in \autoref{fig:fusion_approach}a and our different fusion methods: Naive concatenation as in \autoref{fig:fusion_approach}b left, image concatenation with scaling of encoded hand skeleton image as in \autoref{fig:fusion_approach}b right, multi image input as in \autoref{fig:fusion_approach}c. Best results are listed in \autoref{tab:bestResults}.}
    \label{fig:experiments_2d_boxplot}
    \vspace*{-4mm}
\end{figure}

In \autoref{fig:experiments_2d_boxplot}, we present the results of the baseline experiments with 2D body skeletons for each dataset.
Firstly, it is important to note that the 2D body skeleton baseline results are worse compared to the 3D skeleton baseline results.
This can be attributed to the loss of depth information when using 2D skeletons.

As observed in the previous section on using 3D skeletons, Swin outperforms ResNet on both datasets.
However, as explained in \autoref{sec:datasets} the skeleton-based action recognition problem is very challenging on IKEA ASM due to a differentiation of actions by objects, which are not encoded in skeleton data.
This could explain the smaller improvement in accuracy on IKEA ASM than on ATTACH.

\subsubsection{Fusion of Hand Skeletons with 2D Body Skeletons}
Right to the respective baseline results in \autoref{fig:experiments_2d_boxplot}, we present the results of the fusion experiments with 2D hand and body skeletons.
The comparison between the 2D body skeleton baseline and the fusion approaches reveals a notable improvement in classification performance for both the ATTACH and IKEA ASM datasets.
Thus, the inclusion of hand skeletons in addition to body skeletons emerges as a highly effective strategy to elevate the accuracy of action recognition in assembly applications.
Below, we go into more detail on the results for each dataset individually.

\begin{table}[!b]
\vspace*{-3mm}%
\caption{\label{tab:bestResults}Best results of our experiments. We report the mean class accuracy mAcc and in parentheses the \graytext{top-1 accuracy} for the ATTACH$\,$\cite{attach} and IKEA ASM$\,$\cite{IKEA-wacv2021} datasets.}
\vspace*{-3mm}%
\scriptsize%
\resizebox{\textwidth}{!}{%
\begin{tabular}{llccclcclcc}
\toprule
\multicolumn{1}{c}{} &                   & \multicolumn{6}{c}{ATTACH}                                                                                                                                                                                                        &                   & \multicolumn{2}{c}{IKEA ASM}                                                     \\\rule{0pt}{4mm}%
                     &                   &  \begin{tabular}[c]{@{}c@{}}3D Body\\ {\tiny (Baseline)} \end{tabular}      & \begin{tabular}[c]{@{}c@{}}3D Body\\ 3D Hand\end{tabular} & \begin{tabular}[c]{@{}c@{}}3D Body\\ 2D Hand\end{tabular} &                   &  \begin{tabular}[c]{@{}c@{}}2D Body\\ {\tiny (Baseline)} \end{tabular}       & \begin{tabular}[c]{@{}c@{}}2D Body\\ 2D Hand\end{tabular} &                   &  \begin{tabular}[c]{@{}c@{}}2D Body\\ {\tiny (Baseline)} \end{tabular}                & \begin{tabular}[c]{@{}c@{}}2D Body\\ 2D Hand\end{tabular} \\ \midrule 
ResNet50                                     & \multirow{2}{*}{\phantom{ii}} & 48.2 \graytext{(56.5)} & 48.2 \graytext{(55.7)}                                               & \textbf{48.3} \graytext{(55.9)}                                      & \multirow{2}{*}{\phantom{ii}} & 43.7 \graytext{(52.3)} & \textbf{46.3} \graytext{(54.6)}                                      & \multirow{2}{*}{\phantom{iiiiii}} & \textbf{37.7} \graytext{(70.3)} & 37.2 \graytext{(72.6)}                                               \\\rule{0pt}{4mm}%
SwinV2-T                                     &                   & 52.8 \graytext{(60.3)} & 53.6 \graytext{(61.0)}                                               & \textbf{54.1} \graytext{(61.7)}                                      &                   & 48.1 \graytext{(56.0)} & \textbf{51.5} \graytext{(58.9)}                                      &                   & 39.1 \graytext{(72.6)}          & \textbf{39.9} \graytext{(73.9)}                                      \\ \bottomrule
\end{tabular}
}
\end{table}

\paragraph{ATTACH:}
A closer look on the results on the ATTACH dataset and the comparison with 3D body skeletons reveals that hand skeletons are crucial for achieving improved performance with 2D body skeletons, as indicated by the greater improvement over the corresponding baseline.
This holds true for both Swin and ResNet models, highlighting the significance of hand skeletons in mitigating the loss of depth information when only 2D body skeletons are available.

\paragraph{IKEA ASM:}
The results on the IKEA ASM dataset are less conclusive.
Although the addition of hand skeletons generally leads to better medians and smaller variances, the improvement is not as clear as on the ATTACH dataset.
Specifically, while the Swin and EMSAFormer models show clear improvement with the addition of hand skeletons, the ResNet only shows improvement in median.
One possible explanation for this difference is that predicting hand skeletons on the IKEA ASM dataset is more challenging due to the small size of the hands, which often results in missing hand skeleton estimations.
The attention mechanisms in the Swin transformer may be better suited to handle this issue of jumps in the temporal sequence, while the ResNet struggles with it and therefore processes the information contained in the hand skeletons less effectively.

\section{Conclusion}
\vspace*{-2mm}
Our work demonstrates a successful fusion of hand and body skeletons, which improves assembly action recognition notably.
While hand skeletons contain important information, they are often prone to noise and misinformation due to difficulties in estimation, such as occlusion or object/tool manipulation.
To avoid this issue, our approach specifically handles the importance of the body skeletons to prevent the hand skeletons from dominating the input representation.

Furthermore, our approach demonstrates improved action recognition for two state-of-the-art assembly datasets, not only with 3D body skeletons but also with more commonly available 2D body skeletons.
We have demonstrated a successful approach for preparing hand skeletons for action recognition and provided guidance on the key considerations for successful training with the Swin transformer.
Overall, our work makes an important contribution to the field of action recognition in mobile robotics and collaborative robots.
\vspace*{-2mm}

\bibliographystyle{splncs04}
\def\bibtextsize{\small}
\ifthenelse{\boolean{isarxiv}}{%
    \bibliography{bib}
}{%
    \bibliography{bib_short}
}%
\end{document}